\documentclass[conference]{IEEEtran}
\IEEEoverridecommandlockouts

\usepackage{epsfig}
\usepackage{ifpdf}
\usepackage{textcomp}
\usepackage{xcolor}
\usepackage{subcaption}
\usepackage[colorlinks=true, allcolors=blue]{hyperref}
\usepackage{multicol}

\usepackage{cite}

\usepackage{graphicx}

\usepackage{amsmath, amsfonts, amssymb, amsthm}
\usepackage{bm}

\usepackage{float}

\usepackage{url}

\usepackage{algorithm}
\usepackage{algorithmicx}
\usepackage[noend]{algpseudocode}

\usepackage{graphicx}

\renewcommand{\figurename}{Fig.}

\def\eqalign#1{\null\,\vcenter{\openup\jot\ialign
              {\strut\hfil$\displaystyle{##}$&$\displaystyle{{}##}$
               \hfil\crcr#1\crcr}}\,}

\begin{document}

\title{Unsupervised Machine Learning for Detecting and Locating Human-Made Objects in 3D Point Cloud}

\author{\IEEEauthorblockN{Hong Zhao, Huyunting Huang, Tonglin Zhang, Baijian Yang, Jin Wei-Kocsis, and Songlin Fei}
\IEEEauthorblockA{
\textit{Purdue University}\\
West Lafayette, Indiana \\
\{zhao1211, huan1182, tlzhang, byang, kocsis0, sfei\}@purdue.edu}
}


\IEEEtitleabstractindextext{%
\begin{abstract}
A 3D point cloud is an unstructured, sparse, and irregular dataset, typically collected by airborne LiDAR systems over a geological region. Laser pulses emitted from these systems reflect off objects both on and above the ground, resulting in a dataset containing the longitude, latitude, and elevation of each point, as well as information about the corresponding laser pulse strengths. A widely studied research problem, addressed in many previous works, is ground filtering, which involves partitioning the points into ground and non-ground subsets. This research introduces a novel task: detecting and identifying human-made objects amidst natural tree structures. This task is performed on the subset of non-ground points derived from the ground filtering stage. Marked Point Fields (MPFs) are used as models well-suited to these tasks. The proposed methodology consists of three stages: ground filtering, local information extraction (LIE), and clustering. In the ground filtering stage, a statistical method called One-Sided Regression (OSR) is introduced, addressing the limitations of prior ground filtering methods on uneven terrains. In the LIE stage, unsupervised learning methods are lacking. To mitigate this, a kernel-based method for the Hessian matrix of the MPF is developed. In the clustering stage, the Gaussian Mixture Model (GMM) is applied to the results of the LIE stage to partition the non-ground points into trees and human-made objects. The underlying assumption is that LiDAR points from trees exhibit a three-dimensional distribution, while those from human-made objects follow a two-dimensional distribution. The Hessian matrix of the MPF effectively captures this distinction. Experimental results demonstrate that the proposed ground filtering method outperforms previous techniques, and the LIE method successfully distinguishes between points representing trees and human-made objects.

\end{abstract}

}

\maketitle

\IEEEdisplaynontitleabstractindextext

\IEEEpeerreviewmaketitle

\section{Introduction}
\label{sec:intro}

With recent advances in data collection and storage capabilities, massive amounts of digital data are being generated across a wide ranges of applications, driving the rapid growth of machine learning (ML) and artificial intelligence (AI) tasks and tools in computer science and technologies. Among the various data formats explored by AI/ML approaches, images and 3D point clouds are two typical examples. Compared to 2D image format is well-known, the 3D point cloud format is comparatively less familiar. 
A 3D point cloud consists of unstructured, sparse, and irregular data points that represent one or more 3D shapes or objects from a specific geographical region. In practice, 3D point cloud data are often collected by airborne LiDAR (Light Detection and Ranging) systems, such as those mounted on aircraft, helicopters, or drones, which combine 3D laser scanning technologies
~\cite{shan2018topographic}. 
In addition to the 3D coordinates representing point locations, a 3D point cloud dataset may include additional attributes associated with each point. Due to the fundamental differences between image and 3D point cloud formats, existing ML/AI methods developed for image analysis cannot be directly applied to 3D point clouds. Therefore, specific ML/AI methods tailored for 3D point clouds are required. Since ground truth data are often unavailable, this research focuses on unsupervised ML/AI methods capable of distinguishing between natural and human-made objects. This approach differs from the supervised ML/AI methods studied in PointNet~\cite{qi2017pointnet}, which require ground truth data for training.


\begin{figure}[tb]
\includegraphics[angle=0,width=3.0in]{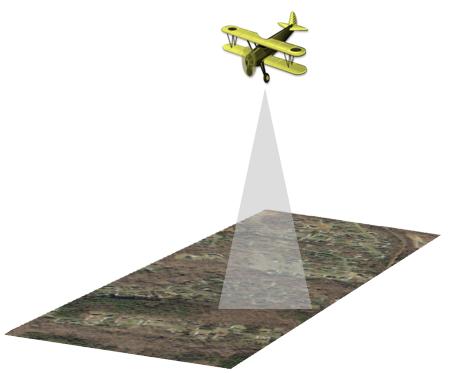}
    \caption{\label{Fig:plane lidar} An aircraft equipped with an airborne Lidar system flying over a geological region.}
\end{figure}

When an airborne Lidar system flies over a geological region, it emits the laser light (infrared, visible, or ultraviolet) toward the ground, with the returning signal received by the Lidar sensors onboard (Figure~\ref{Fig:plane lidar}). The two types of airborne sensors are topographic and bathymetric. A topographic LiDAR sensor provides surface models for landscapes. Examples include forestry, hydrology, geomorphology, urban planning, and landscape ecology. A bathymetric LiDAR sensor collects elevation and water depth simultaneously for surveys of land-water interfaces. Bathymetric information is important near coastlines, harbors, shores, and banks. This information is also useful in locating objects on the ocean floor.

Locations of 3D LiDAR points are represented by coordinates in $\mathbb{R}^3$, corresponding to their longitude, latitude, and elevation values. A LiDAR point may also have associated strength variables for the laser pulse that generated the LiDAR point. Other associated variables might include the classifications of the points that have reflected the laser pulse. However, due to the poor performance of classification models used by the sensors, classifications are often missing or incorrect in 3D point clouds. Therefore, it is inappropriate to use these variables in the analysis of LiDAR data. The typical 3D point cloud dataset generally contains only the 3D coordinates along with the corresponding laser pulse strengths of the points.
\begin{figure}[tb]
     \centering
     \begin{subfigure}[hb]{0.18\textwidth}
         \centering
         \includegraphics[width=\textwidth]{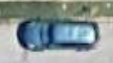} \\
         \includegraphics[width=\textwidth]{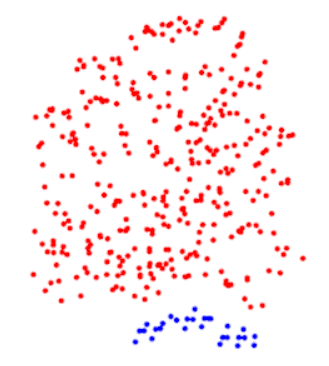}
         \caption{}
     \end{subfigure}
     \begin{subfigure}[hb]{0.18\textwidth}
         \centering
         \includegraphics[width=\textwidth]{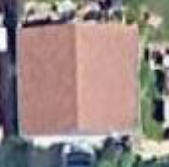} \\
         \includegraphics[width=\textwidth]{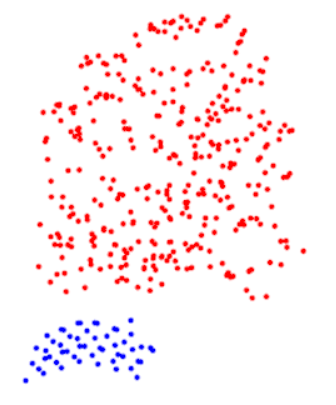}
         \caption{}
     \end{subfigure}
     \caption{\label{fig:tree, car, and hourse}3D point cloud for (a) a car beneath a tree; (b) a house beneath a tree}
\end{figure}

Laser pulses emitted from a LiDAR system reflect off objects on and above the ground. These objects may include vegetation, water surfaces, buildings, bridges, roads, and vehicles. In many cases, vegetation, defined as an assemblage of plant species, covers most of the study region, where human-made objects may also be present. A key research question is detecting and locating human-made objects within a vegetated area. Due to the variety of human-made objects and the lack of ground truth data, it is inappropriate to employ a supervised learning approach to address this research question. Instead, we devise an unsupervised learning method for this task.

Marked Point Fields (MPFs) are the most appropriate ML/AI models for 3D point cloud data. From a statistical perspective, an MPF is a collection of random locations (i.e., points) with corresponding measurements (i.e., marks) collected in Euclidean space. It becomes a Marked Point Process (MPP) if the Euclidean space is one-dimensional. MPFs are often used to model natural hazard events, where points represent the locations and marks represent the magnitudes. Examples include earthquakes \cite{ogata1998space,zhang2017independence,zhuang2002stochastic} and forest wildfires \cite{peng2005space,schoenberg2004testing,zhang2014kolmogorov}. For the 3D point cloud, points represent the reflection coordinates, and marks represent the laser pulse strengths. An unsupervised ML/AI method based on MPFs partitions the points into trees or human-made objects.

We propose our method based on the differences between the geometric shapes of trees and human-made objects. We illustrate our concept in Figure~\ref{fig:tree, car, and hourse}. The laser light can be returned from the points on the surface or in the interior of the trees (e.g., leaves or trunks), but it can only be returned from the points on the surface of human-made objects. Because the interior points are distributed three-dimensionally, the geometric shapes of the points for the trees are three-dimensional. Because the body surface of human-made objects is usually smooth, the geometric shapes of the points for the human-made objects are two-dimensional. Based on this property, we devise an unsupervised ML/AI method to detect the existence of a smooth surface in a geological region. We partition the points into two subsets: one is three-dimensional and the other is two-dimensional. We claim human-made subjects are not contained in the study region if the two-dimensional subset is empty. We locate the human-made objects when the two-dimensional subset is nonempty.

\begin{figure}[tb]
     \centering
     \begin{subfigure}[b]{0.22\textwidth}
         \centering
         \includegraphics[width=\textwidth]{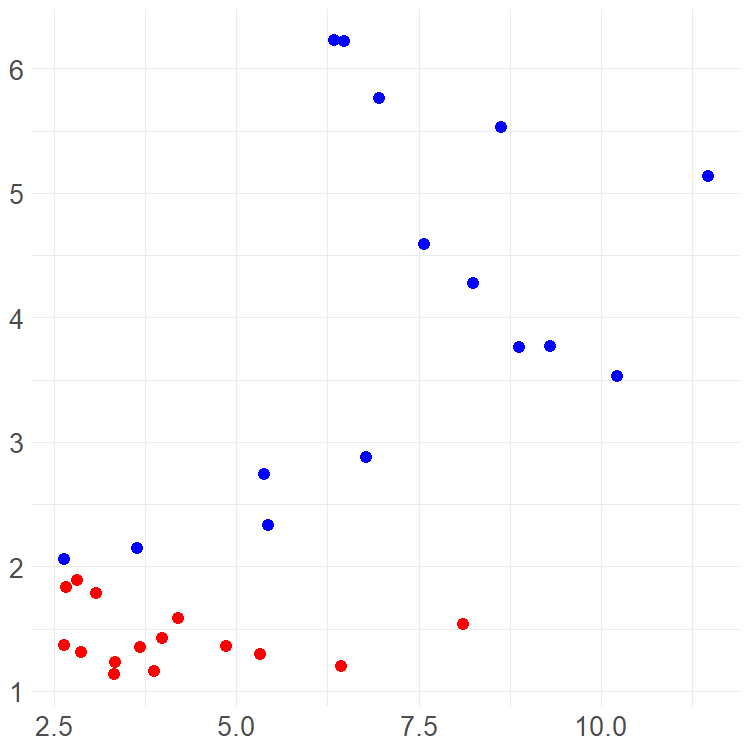}
         \caption{}
     \end{subfigure}
     \begin{subfigure}[b]{0.22\textwidth}
         \centering
         \includegraphics[width=\textwidth]{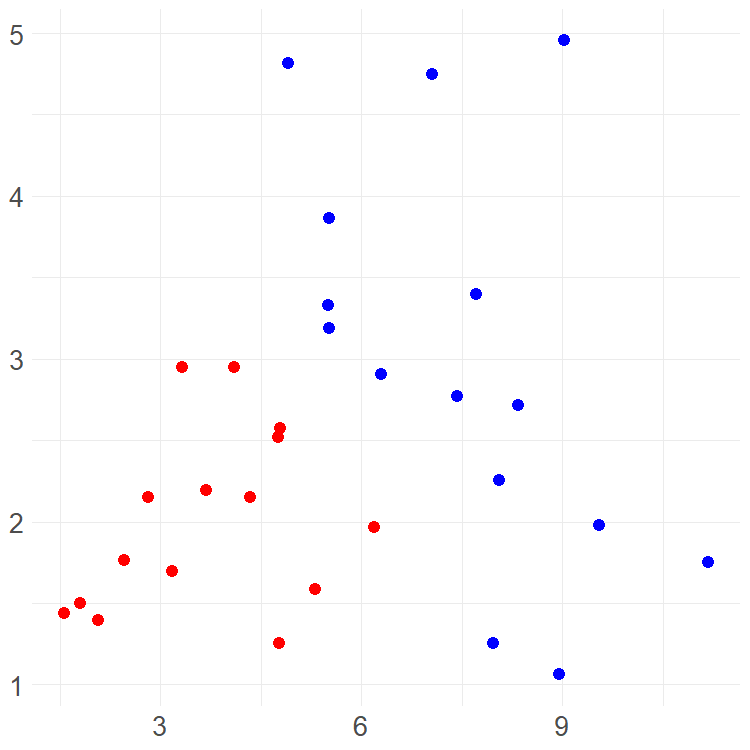} 
         \caption{}
     \end{subfigure}
     \caption{\label{fig:log of the proportion of the smallest and the second eigenvalus}Scatter plots for the logarithms of the ratios of localized eigenvalues for selected points in Figure 2}
\end{figure}

We devise local information extraction (LIE) for the 3D point cloud. Previous methods for the MPFs rely on a concept called intensities~\cite{daley2003introduction}. Recent work shows that the gradient vector of an intensity function can capture more useful information about local variations~\cite{zhang2017gradient}. Following this approach, we investigate the Hessian matrix of intensities. We find that it is an appropriate measure to determine the dimension of the points locally. From a mathematical perspective, any two-dimensional surface can be locally approximated by a plane. In this case, the Hessian matrix of the intensity function is nearly degenerate, \textit{i.e.} the least absolute eigenvalue of the Hessian matrix approaches to $0$. In contrast, a 3D geometric shape does not exhibit this property. Its least absolute eigenvalue is significantly greater than $0$. Leveraging this property, we compute the eigenvalues of the Hessian matrix for the points in Figure~\ref{fig:tree, car, and hourse}. We plot the logarithms of the ratios of the least and the second least eigenvalue squares, resulting in Figure~\ref{fig:log of the proportion of the smallest and the second eigenvalus}. We find that the points from the trees and human-made objects are well-separated. 

Ground filtering (or filtering) is essential and serves as the first step in many ML/AI methods. The task is to separate ground and nonground points from the 3D point cloud. The result is critical for object recognition, classification, and feature extraction. However, separating ground and nonground returns has become challenging due to the mixture of ground and vegetation returns. This challenge persists due to the complexity of the input data and the real-time requirements. 
We recognize that the elevations of ground points are lower than those of nonground points on horizontal terrain. We extend this observation and present an innovation statistical ground filtering method, denoted as One-Sided Regression (OSR), to address the ground filtering problem. 
This is based on the fact that nonground points are outliers in a statistical model and the residuals of the outliers can only be positive.
One advantage of OSR is that it effectively handles the difficulties posed by uneven terrains. Based on this property, we derive a mathematical expression for the ground surface. Using our statistical ground filtering method, we successfully separate the ground and nonground points. We then detect trees and human-made objects based on the subset of nonground points provided by the ground filtering process.

The rest of the article is organized as follows. In Section~\ref{sec:related work}, we review the related work. In Section~\ref{sec:method}, we introduce our method. In Section~\ref{sec:experiment}, we present our experimental results. Finally, in Section~\ref{sec:conclusion}, we conclude this work.

\section{Related Work}
\label{sec:related work}

\begin{figure}[tb]
    \centering
    \begin{subfigure}[b]{0.22\textwidth}
        \centering
        \includegraphics[width=\textwidth]{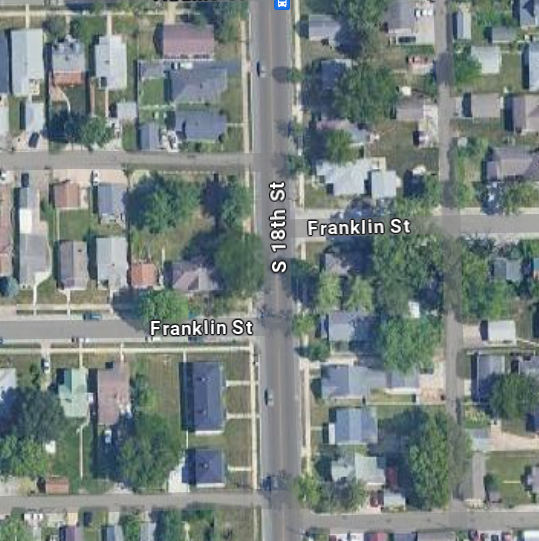}
        \caption{Satellite map}
    \end{subfigure}
    \begin{subfigure}[b]{0.22\textwidth}
         \centering
         \includegraphics[width=\textwidth]{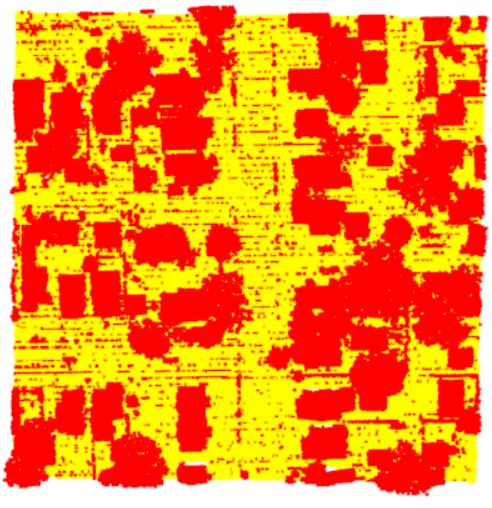} 
         \caption{Original labels}
     \end{subfigure}
     \caption{\label{fig:satellite map city} (a) Satellite map of an airborne LiDAR dataset from a region in West Lafayette, Indiana; (b) original labels for {\it ground} points (yellow) and {\it nonground} points (red).}
\end{figure}

The three stages of our method are ground filtering, LIE, and clustering. Ground filtering continues to be a challenging task, even after two decades of research. Its results significantly impact object recognition and clustering for 3D point clouds. LIE extracts useful local features from the subset of nonground points, with the goal of ensuring that existing ML/AI methods can be utilized. LIE is applied to the output of the ground filtering stage and has not been well-studied in the literature. Clustering, a typical unsupervised learning approach, is then applied to the results of the LIE stage. We review previous work corresponding to these three stages.

Ground filtering methods are classified into slope-based, mathematical morphology-based, and surface-based approaches. Slope-based methods are developed under the assumption that variations in the slopes of terrains can be ignored locally, while variations in the slopes of nonground objects (e.g., trees and buildings) cannot. Based on this assumption, a sloped-based filtering algorithm is proposed~\cite{vosselman2000slope}. To enhance computational efficiency, sloped-based filtering algorithms along one specific scan line~\cite{shan2005urban} and a few scan lines~\cite{meng2009multi} were developed. Due to the challenge of selecting a slope threshold applicable to various terrain types, slope-based filtering algorithms with various automatic thresholds have been proposed~\cite{sithole2001filtering,streutker2011slope,susaki2012adaptive,wan2018simple}. A concern, however, is that slope-based algorithms may not work well in complicated terrains~\cite{liu2008airborne}.

\begin{figure}[tb]
    \centering
    \centering
    \begin{subfigure}[b]{0.22\textwidth}
        \centering
        \includegraphics[width=\textwidth]{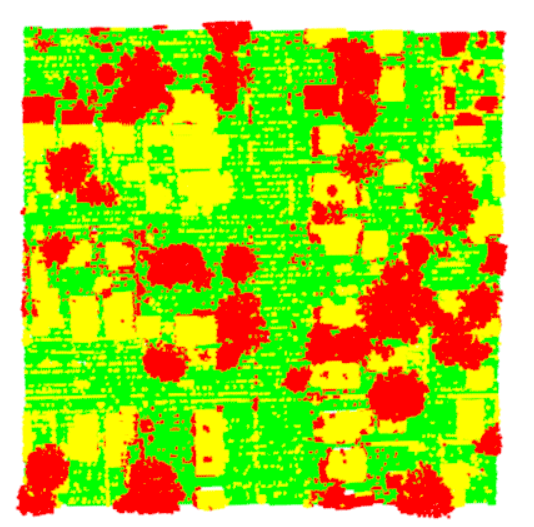}
        \caption{Trees vs human-made objects for nonground points}
    \end{subfigure}
    \begin{subfigure}[b]{0.21\textwidth}
        \centering
        \includegraphics[width=\textwidth]{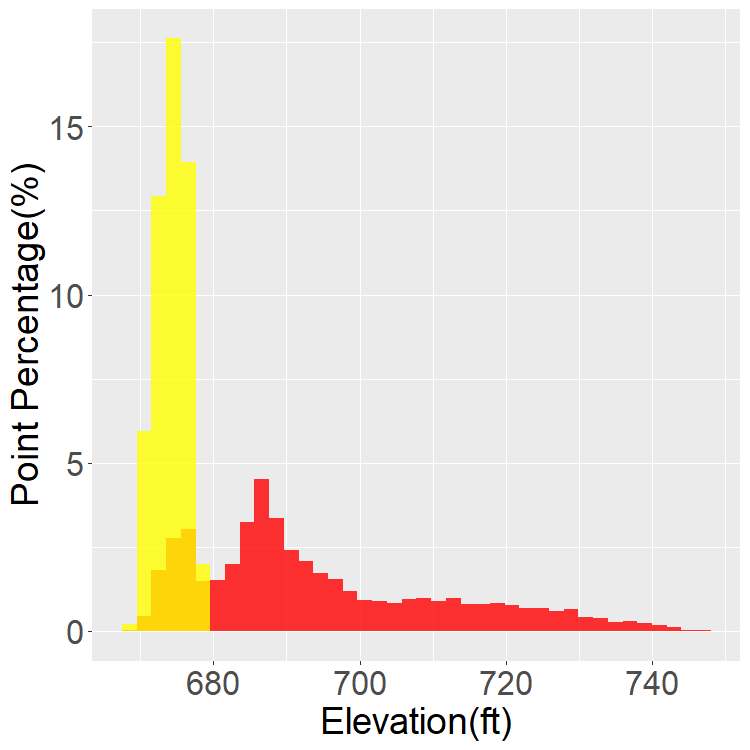}
        \caption{Histogram of elevations}
    \end{subfigure}
    \caption{\label{fig:Cluster without ground filtering}(a) Unsupervised ML/AL for human-made objects (yellow) and trees (red) derived by  combining our proposed LIE method with the previous GMM method (the clustering stage) for the {\it nonground} points in Figure~\ref{fig:satellite map city}(b); (b) Histogram showing the percentage of the elevations for {\it ground} (yellow) and {\it nonground} (red) points in Figure~\ref{fig:satellite map city}(b).
    }
\end{figure}

Mathematical morphology-based methods use the moving window approach to separate ground and nonground points. To address the concern related to the choice of window size~\cite{sithole2005filtering}, a progressive morphological filter (PMF) method with varied window sizes is proposed~\cite{zhang2003progressive}. Since many mathematical morphology-based methods treat terrain slope as a constant, the PMF has been modified to account for local terrain topography through a set of tunable parameters~\cite{chen2007filtering}. Mathematical morphology-based methods are conceptually simple and easy to implement, but they require additional prior knowledge to determine suitable window sizes for local operators~\cite{mongus2012parameter}.

Surface-based methods approximate the ground surface. Early examples include the adaptive triangulated irregular network filtering algorithm~\cite{axelsson2000generation}, the weighted least squares interpolation~\cite{kraus1998determination}, and the hierarchical interpolation~\cite{pedregosa2011scikit}. These methods perform well on flat terrain but struggle on steep terrain. To address this issue, the multi-resolution hierarchical filtering~\cite{chen2013multiresolution,su2015new,hui2016improved} and the active shape model~\cite{elmqvist2000automatic} were proposed. However, these methods may still perform poorly in areas with high terrain variability~\cite{guan2014generation}. To further tackle this problem, the cloth simulation filtering (CSF)~\cite{zhang2016easy} and the multiscale curvature classification (MCC)~\cite{evans2007multiscale} methods were introduced.

The MPF approach will be used in the LIE stage. The goal is to extract useful local features for human-made object detection and location based on the nonground points provided by ground filtering. In the literature, an MPF is treated as a random pattern of points developed in an Euclidean space. MPFs are typically described by intensity functions~\cite{diggle2013statistical}. Various well-known tools have been developed under an important concept called stationarity. Examples include the K-function~\cite{ripley1976second}, the L-function~\cite{besag1977contribution}, and the pair correlation function~\cite{stoyan1996estimating}. Due to concerns regarding stationarity, recent work has focused on nonstationarity~\cite{moller2007modern}. Examples include the second-order intensity-reweighted stationarity~\cite{baddeley2000non} and substationarity~\cite{zhang2019substationarity} methods. These methods, however, cannot be applied in the LIE stage because they were developed under stationarity or its extended version. In this research, we discard stationarity and devise a new LIE method (Section~\ref{sec:method}). 
 
Clustering is applied to the local features provided by the LIE. It is a popular unsupervised ML/AL approach in the literature. Clustering is typically performed using the well-known Gaussian mixture model (GMM)~\cite{loffler2021optimality} or the $k$-means~\cite{huang2023improved} methods. The clustering stage is relatively straightforward compared to the ground filtering and LIE stages. 

We investigated the foundation of our method. We downloaded a 3D point dataset for a $500\times 500({\rm ft}^2)$ region with a $0.35/{\rm ft}^2$ point density in Lafayette, Indiana, from the website at {\it https://lidar.digitalforestry.org/} (Figure~\ref{fig:satellite map city}(a)). The dataset contained the longitude, latitude, and elevation values for the points and a few marks associated with the points. An important mark was the classification of ground or nonground (labeled as {\it unassigned}) for points derived by several slope-based methods (Figure~\ref{fig:satellite map city}(b)). We extracted local features for the nonground points using our proposed OSR method. We then applied existing GMM method for clustering the nonground points into a subset for trees and another subset for human-made objects (Figure~\ref{fig:Cluster without ground filtering}(a)). However, we found many classification errors (e.g., the upper right corner, the upper left corner, the middle of the left, etc). One concern was the ground filtering methods adopted by the website. The slope-based methods misclassified too many ground points as nonground points (i.e., red points with low elevations in Figure~\ref{fig:Cluster without ground filtering}(b)).

\section{Method} 
\label{sec:method}

We face two issues in our methodological development. The previous ground filtering methods may not work well. There is a lack of LIE methods for local features. We can apply the previous GMM and $k$-means for the clustering stage, implying that it is not an issue. Therefore, we only propose our methods for the ground filtering and LIE stages. 

\subsection{Statistical Ground Filtering}
\label{secsub:statistical ground filtering}

Our statistical ground filtering method is called one-sided regression (OSR). It estimates the ground surface and separates the ground/nonground points. We treat the ground points as observations from the ground surface and nonground points as the outliers. Locally, the ground surface can be approximated by a central plane. The distance between a ground point and the central plane is small but the distance of a non-ground point is not. The OSR estimates the mathematical formulation of the central plane but not the ground surface. It uses the property that the nonground points are above the ground surface. Compared with the previous methods (i.e., sloped-based, mathematical morphology-based, or surface-based), the proposed OSR does require the ground surface to be gradually varied, implying that it can be applied to uneven terrains. 

Let ${\bm p}_i=(x_i,y_i,z_i)^\top$, $i=1,\dots,n$, be the coordinates of the $i$th point in the 3D point cloud, where $n$ is the number of points. Suppose the ground surface is close to a central plane expressed as $z=\beta_0+\beta_1x+\beta_2y$ with the variations from the central plane controlled by random errors. The variance of the random effect provides the magnitude of the unevenness of the terrain. 

We propose our statistical model for the points as 
\begin{equation}
\label{eq:statistical model for ground/nonground points}
    z_i=\beta_0+\beta_1x_i+\beta_2y_i+\delta_i+\epsilon_i,
\end{equation}
where $\epsilon_i\stackrel{iid}{\sim}{\mathcal N}(0,\sigma^2)={\mathcal N}(0,\phi)$ is the random error and $\delta_i$ is an index variable for ground or nonground points. We use $\epsilon_i$ to describe the topography of the study region. A large value of $\phi$ indicates that the magnitude of the unevenness of the terrain is large or vice versa. The OSR method does not suffer from the challenges caused by uneven terrains. 

We set $\delta_i>0$ for a nonground point or $\delta_i=0$ for a ground point. The set of nonground points is ${\mathcal A}=\{i:\delta_i>0\}$. The set of ground points is the complementary set of ${\mathcal A}$ as ${\mathcal A}^c=\{i:\delta_i=0\}$. Given ${\mathcal A}$, the log likelihood function (i.e., the objective function) of our OSR for~\eqref{eq:statistical model for ground/nonground points} is
\begin{equation}
\label{eq:objective function of the OSR}
\eqalign{
\ell({\bm\beta},\phi|{\mathcal A})=&\sum_{i\not\in{\mathcal A}} \left[-\log(2\pi\phi)-{1\over 2\phi} (z_i-\eta_i)^2\right],\cr
}
\end{equation}
where ${\bm\beta}=(\beta_0,\beta_1,\beta_2)^\top$. The corresponding estimators of ${\bm\beta}$ and $\phi$ are derived by
\begin{equation}
\label{eq:estimator OSR}
(\hat{\bm\beta}_{\mathcal A},\hat\phi_{\mathcal A})=\mathop{\arg\!\max}_{{\bm\beta},\phi}\ell({\bm\beta},\phi|{\mathcal A}).
\end{equation}
Because ${\mathcal A}$ is unknown, we estimate ${\mathcal A}$ before the implementation of~\eqref{eq:objective function of the OSR} and~\eqref{eq:estimator OSR}. 

The distance between a ground point and the central plane depends on the random error only. The distance of a nonground point is also affected by $\delta_i$. Let 
\begin{equation}
\label{eq:residual of the model for ground/nonground points}
e_i=z_i-\hat z_i
\end{equation}
be the residual of point $i$, where $\hat z_i=\hat\beta_0+\hat\beta_1x_i+\hat\beta_2y_i$ is the fitted value of $z_i$, and $\hat\beta_0$, $\hat\beta_1$, and $\hat\beta_2$ are the estimates of $\beta_0$, $\beta_1$, and $\beta_2$ derived by an estimation method, respectively. Because $\delta_i$ cannot be negative in~\eqref{eq:statistical model for ground/nonground points}, point $i$ is classified as a ground point if $e_i\le 0$. It may be classified as a nonground point if $e_i>0$, depending on whether $e_i$ is large or not. The OSR used this property to estimate the model parameters of~\eqref{eq:statistical model for ground/nonground points} (i.e., $\beta_0$, $\beta_1$, $\beta_2$, $\phi$, and ${\mathcal A}$).

The computing of the OSR is proposed based on the left-sided half-normal distributions. A left-sided half-normal distribution is derived if the positive part of a mean zero normal distribution is set to be $0$. If the true distribution is $U\sim{\mathcal N}(0,\sigma^2)$, then the left-sided half normal distribution $U^*$ is derived by setting $U^*=UI_{U\le 0}$, where $I_{U\le0}=1$ if $U\le 0$ or $I_{U\le0}=0$ otherwise. Based on this approach, we set $z_i$ to be its observed value if $\delta_i+\epsilon_i\le 0$ or $z_i=\eta_i=\beta_0+\beta_1x_i+\beta_2 y_i$ otherwise, leading to 

We numerically optimize the right-hand side of~\eqref{eq:estimator OSR}. The main issue is to estimate ${\mathcal A}$ denoted as $\hat{\mathcal A}$. We devise an iterative algorithm to fulfill the task. Let ${\bm\beta}^{(t)}=(\beta_0^{(t)},\beta_1^{(t)},\beta_2^{(t)})^\top$ be the iterative vector of $\hat{\bm\beta}$ in the $t$th iteration. The residuals of the model based on ${\bm\beta}^{(t)}$ are $e_i^{(t)}=z_i-z_i^{(t)}$, $i=1,\dots,n$, where $z_i^{(t)}=\beta_0^{(t)}+\beta_1^{(t)}x_i+\beta_2^{(t)}y_i$ is the $t$th predicted value for $z_i$. The $t$th iteration claims point $i$ a ground point (i.e., $i\not\in{\mathcal A}$) if $e_i^{(t)}\le 0$, leading to the $t$th iterated value of $\hat\phi$ as
\begin{equation}
\label{eq:iterative value of phi}
\phi^{(t)}={\sum_{i: e_i^{(t)}\le 0} \{e_i^{(t)}\}^2\over\sum_{i=1}^n I_{e_i^{(t)}\le 0}}.
\end{equation}
By the work of~\cite{donoho2004higher}, we estimate ${\mathcal A}$ in the $t$th iteration as
\begin{equation}
\label{eq:estimates of the outliers}
{\mathcal A}^{(t)}=\{i: e_i^{(t)}>\sqrt{2\phi^{(t)}\log{n}}\}.
\end{equation}
We compute the right-hand side of~\eqref{eq:estimator OSR} by putting this in~\eqref{eq:objective function of the OSR}. The next iterated value of $\hat{\bm\beta}$ is computed by the standard least squares approach based on $\{{\mathcal A}^{(t)}\}^{c}$ as
\begin{equation}
\label{eq:the next iterated value of hat beta}
{\bm\beta}^{(t+1)}=(\{{\bf C}^{(t)}\}^\top\{{\bf C}^{(t)}\})^{-1}\{{\bf C}^{(t)}\}^\top{\bm z}^{(t)},
\end{equation}
where the $i$ row of matrix ${\bf C}^{(t)}$ is ${\bm c}_i=(1,x_i,y_i)$ and the $i$th component of ${\bm z}^{(t)}$ is $z_i$ for all $i\not\in{\mathcal A}^{(t)}$. The derivation of ${\bf C}^{(t)}$ does not use any $i\in{\mathcal A}^{(t)}$. The formulation for the next $\hat{\bm\beta}$ given by~\eqref{eq:the next iterated value of hat beta} can be implemented only when ${\mathcal A}^{(t)}$ given by~\eqref{eq:estimates of the outliers} has been derived. By~\eqref{eq:iterative value of phi},~\eqref{eq:estimates of the outliers}, and~\eqref{eq:the next iterated value of hat beta}, we obtain the iterations. To start the iterations, we choose the initial ${\mathcal A}=\emptyset$ for the nonground points, leading to ${\mathcal A}^c=\{1,\dots,n\}$ for the ground points. At the end of the iterations, we obtain $\hat{\bm\beta}$, $\hat\phi$, and $\hat{\mathcal A}$, the estimators of ${\bm\beta}$, $\phi$, and ${\mathcal A}$, respectively. We treat $\hat{\mathcal A}$ as the estimator of the subset of the nonground points and the complementary set of $\hat{\mathcal A}$ as the estimator of that of ground points. We obtain Algorithm~\ref{alg:one-sided regression algorithm}.

\begin{algorithm}[tb]
\caption{The Proposed OSR for Ground Filtering}
\label{alg:one-sided regression algorithm}
\begin{flushleft}
\textbf{Input}: Points $\{(x_i,y_i,z_i):i=1,\dots,n\}$\\
\textbf{Output}: $\hat{\bm\beta}$, $\hat\phi$, and $\hat{\mathcal A}$
\end{flushleft}
\begin{algorithmic}[1] 
\Statex{Initialization}
\State{Set ${\mathcal A}^{(0)}=\emptyset$ in~\eqref{eq:objective function of the OSR}}
\State{Compute ${\bm\beta}^{(1)}$ by the ordinary least squares}
\Statex{Iteration}
\State{Set $e_i^{(t)}=z_i-\beta_0^{(t)}-\beta_1^{(t)}x_i-\beta_2^{(t)}y_i$ using ${\bm\beta}^{(t)}$}
\State{Compute $\phi^{(t)}$ by~\eqref{eq:iterative value of phi} and ${\mathcal A}^{(t)}$ by~\eqref{eq:estimates of the outliers}}
\State{Compute the next ${\bm\beta}^{(t+1)}$ by~\eqref{eq:the next iterated value of hat beta} and goes to Step 3 until convergence}
\Statex{Output}
\State{The final $z=\beta_0^{(t)}+\beta_1^{(t)}x+\beta_2^{(t)}y$ as the central plane}
\State{The final $\sqrt{\phi^{(t)}}$ as the unevenness of the terrain}
\State{The final ${\mathcal A}^{(t)}$ as the subset of nonground points and its complementary as the subset of ground points}
\end{algorithmic}
\end{algorithm}

Algorithm~\ref{alg:one-sided regression algorithm} does not provide the mathematical formulation of the ground surface. It uses a central plane to approximate the ground surface. The variability is measured by $\hat\phi^{1/2}$. The ground surface is almost identical to the central plane for flat terrains, where $\hat\phi$ is small. The ground surface fluctuates the central planes for uneven terrains, where $\hat\phi^{1/2}$ is large. The determination of the subset of nonground points depends on $\hat\phi$. It sets point $i$ as nonground if randomness cannot interpret the difference. 

\begin{figure}[tb]
\includegraphics[angle=270,width=3.4in]{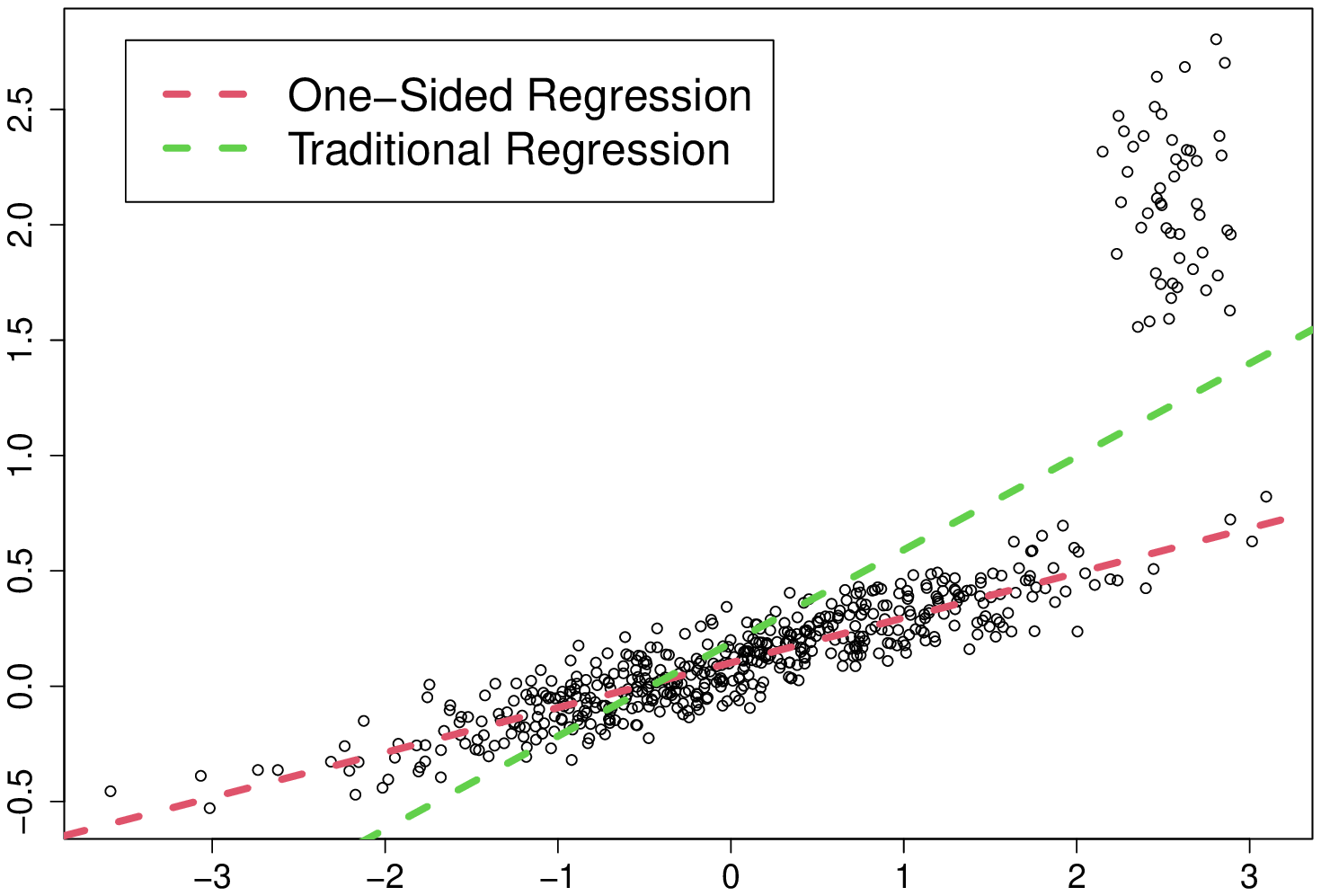}
    \caption{\label{fig:one-sided regression} The central planes for the ground surfaces, derived using the one-sided regression (OSR, red) and traditional regression (green) methods, respectively.}
\end{figure}

We compare the proposed OSR with the traditional regression for~\eqref{eq:statistical model for ground/nonground points}. The traditional regression is a two-sided method because it uses the positive and negative residuals to estimate $\phi$. We simulated an unevenly inclined ground surface with nonground objects above the ground surface. We estimated the central plane of the ground surface by our OSR and the previous traditional regression methods, leading to Figure~\ref{fig:one-sided regression}. The result showed that the proposed OSR was better. 

\subsection{Local Information Extraction (LIE)}
\label{subsec:local information extraction (lie)}

The LIE stage is applied to the subset of nonground points provided by the ground filtering stage. It extracts local features for the nonground points. Points $i\not\in\hat{\mathcal A}$ are ignored. It is different from the traditional methods for the MPFs. They focus on the stationarity of the mean and covariance functions while we focus on local variations. 

Let $N(\cdot)$ be an MPF defined on domain ${\mathcal D}\subset\mathbb{R}^3$ with $N(B)=\#\{i\in{\mathcal A}: {\bm p}_i\in B\}$ representing the number of nonground points in $B\subseteq{\mathcal D}$ (assume marks are ignored). The mathematical method for MPFs assumes $N(B)$ is random. The distribution is determined by its intensity functions. Let $\lambda({\bf s})$ and $\lambda({\bm s}_1,{\bm s}_2)$ with ${\bm s},{\bm s}_1,{\bm s}_2\in{\mathcal D}$ be the first and second-order intensity function of $N(\cdot)$. The first-order intensity satisfies 
\begin{equation}
\label{eq:first order intensity function point field}
{\rm E}[N(B)]=\int_B \lambda({\bm s})d{\bf s}
\end{equation}
for any measurable $B\subseteq{\mathcal D}$. The second-order intensity function satisfies 
${\rm cov}[N(B_1),N(B_2)]=\int_{B_1}\int_{B_2}[g({\bf s}_1,{\bf s}_2)-1]\lambda({\bm s}_1)\lambda({\bm s}_2)d{\bm s}_2d{\bm s}_1+ \int_{B_1\cup B_2}\lambda({\bm s})d{\bf s}$ for any measurable $B_1,B_2\subseteq{\mathcal D}$, where $g({\bm s}_1,{\bm s}_2)=\lambda_2({\bm s}_1,{\bm s}_2)/[\lambda({\bf s}_1)\lambda({\bm s}_2)]$ is the pair correlation function. 

The previous methods for MPFs are developed under a concept called stationarity, the properties of $\lambda({\bm s})$ and $g({\bm s}_1,{\bm s}_2)$. It is said that the MPF $N(\cdot)$ is first-order stationary if $\lambda({\bm s})=\lambda({\bm s}+{\bm u})$ for any ${\bm u}$ in the domain. It is said that $N(\cdot)$ is second-order intensity-reweighted stationary if the pair correlation function can be expressed as $g({\bm s}_1-{\bm s}_2)$. 

Stationarity is irrelevant to our research task. The previous MPF methods are inappropriate for detecting human-made objects in 3D point clouds. We find it appropriate to use the Hessian matrix of the MPF to fulfill our research. Our idea is motivated by Figure~\ref{fig:tree, car, and hourse}. The points on the surface of human-made objects are distributed two-dimensionally. The points on the surface or in the interior of trees are distributed three-dimensionally. The Hessian matrix of the distribution of $N(\cdot)$ can detect the difference. We find that using the Hessian matrix of the first-order intensity is enough. 

Let $\dot\lambda({\bm p})=(\partial\lambda({\bm p})/\partial x,\partial\lambda({\bm p})/\partial y,\partial\lambda({\bm p})/\partial z)^\top$ be the gradient vector and
\begin{equation}
\label{eq:hessian matrix}
H({\bm p})=\ddot\lambda({\bm p})=\left(\begin{array}{ccc} {\partial^2\lambda({\bm p})\over\partial x^2} &  {\partial^2\lambda({\bm p})\over\partial x\partial y} &  {\partial^2\lambda({\bm p})\over\partial x\partial z} \cr 
{\partial^2\lambda({\bm p})\over\partial y\partial x} &  {\partial^2\lambda({\bm p})\over\partial y^2} &  {\partial^2\lambda({\bm p})\over\partial y\partial z} \cr 
{\partial^2\lambda({\bm p})\over\partial z\partial x} &  {\partial^2\lambda({\bm p})\over\partial z\partial y} &  {\partial^2\lambda({\bm p})\over\partial z^2} \cr 
\end{array} \right)
\end{equation}
be the Hessian matrix of $\lambda({\bm p})$, where ${\bm p}=(x,y,z)^\top\in{\mathcal D}$ can be arbitrary. Then, $H({\bm p})$ is almost singular if ${\bm p}$ belongs to human-made objects. The three eigenvalues of $H({\bm p})$ are far away from each other with the least almost $0$. If ${\bm p}$ belongs to trees, then the three eigenvalues of $H({\bm p})$ are close to each other. We can detect whether ${\bm p}$ is from human-made objects or trees by examining whether the least absolute eigenvalue of $H({\bm p})$ is close to $0$. Because $\lambda({\bm p})$ is unknown, we estimate $H({\bm p})$ at the nonground points. 

We devise a kernel method to estimate $H({\bm p}_i)$, $i\in{\mathcal A}$. The traditional kernel method estimates $\lambda({\bm p})$ for ${\bm p}\in{\mathcal D}$ as
\begin{equation}
\label{eq:kernel estimator of the first-order intensity}
\hat\lambda({\bm p})=\sum_{i\in{\mathcal A}} K_{\bm h}({\bm p}_i-{\bm p}),
\end{equation}
where $K_{\bm h}({\bm s})$ is a kernel function at ${\bm s}$ and ${\bm h}=(h_x,h_y,h_z)^\top$ is a bandwidth vector. 
We estimate $H({\bm p})$ by
\begin{equation}
\label{eq:kernel estimator of the Hessian matrix}
\eqalign{
\hat H({\bm p})=&\sum_{i\in{\mathcal A}} \ddot K_{\bm h}({\bm p}_i-{\bm p})\cr=&\left(\begin{array}{ccc} \hat H_{xx}({\bm p}) &  \hat H_{xy}({\bm p}) & \hat H_{xz}({\bm p}) \cr
\hat H_{yx}({\bm p}) &  \hat H_{yy}({\bm p}) & \hat H_{yz}({\bm p}) \cr
\hat H_{zx}({\bm p}) &  \hat H_{zy}({\bm p}) & \hat H_{zz}({\bm p}) \cr
\end{array} \right),
}\end{equation}
where $\ddot K_{\bm h}({\bm s})$ is the Hessian matrix of the kernel function at ${\bm s}$. The properties of~\eqref{eq:kernel estimator of the first-order intensity} and~\eqref{eq:kernel estimator of the Hessian matrix} depend on the kernel function. We suggest the Gaussian kernel as
\begin{equation}
\label{eq:3D gaussian kernel}
\eqalign{
K_{\bm h}&({\bm p}_i-{\bm p})=[(2\pi)^{3/2}h_xh_yh_z]^{-1}\cr
&\exp\left[-{(x_i-x)^2\over 2h_x^2}-{(y_i-y)^2\over 2h_y^2}-{(z_i-z)^2\over 2h_z^2
}\right].
}
\end{equation}
Putting~\eqref{eq:3D gaussian kernel} into~\eqref{eq:kernel estimator of the Hessian matrix}, we obtain $\hat H_{xx}({\bm p})=\sum_{i\in{\mathcal A}} K_{\bm h}({\bm p}_i-{\bm p})[(x_i-x)^2-h_x^2]/h_x^4$ and $\hat H_{xy}({\bm p})=\sum_{i\in{\mathcal A}} K_{\bm h}({\bm p}_i-{\bm p})(x_i-x)(y_i-y)/(h_x^2h_y^2)$ with the remaining entries expressed analogously. 

It is important to choose an appropriate ${\bm h}$ in the implementation of~\eqref{eq:kernel estimator of the Hessian matrix}. This is called the bandwidth selection problem in the statistical literature. Previous approaches assume $h_x=h_y=h_z$ to select the optimal bandwidth. We set $h_x=h_y\not=h_z$ in our method, implying that we need two bandwidth values. The result is usually reliable in bandwidth values. The only issue is that $h_x$ and $h_z$ cannot be too small in applications. 

Note that ${\mathcal A}$ is unknown. We apply~\eqref{eq:kernel estimator of the Hessian matrix} for every $i\in\hat{\mathcal A}$, where $\hat{\mathcal A}$ is provided by a ground filtering method. The ground filtering and the LIE stages of our method are independently developed. Formulation~\eqref{eq:kernel estimator of the Hessian matrix} can be combined with arbitrary ground filtering methods. All eigenvalues of $\hat H({\bm p})$ are real because it is symmetric. We choose 
\begin{equation}
\label{eq:choice of local variables eigenvalue}
v_i=-\log{\lambda_1^2[\hat H({\bm p}_i)]\over \sum_{j=1}^3 \lambda_j^2[\hat H({\bm p}_i)]}
\end{equation}
 as the value of the local feature for point $i$ with $i\in\hat{\mathcal A}$, where $\lambda_j[\hat H({\bm p}_i)]$ represents the $j$th least eigenvalue of $\hat H({\bm p}_i)$ in absolute values. We expect $v_i$ to be large if ${\bm p}_i$ is a point from human-made objects or small otherwise.

The derivation of our clustering stage is straightforward. We use $v_i$ to detect and locate human-made objects in the 3D point cloud. We apply $v_i$ for all $i\in\hat{\mathcal A}$ to a clustering method, where the associated marks may also be involved. We partition $\hat{\mathcal A}$ into two subsets. The subset of trees is composed of the points with low values of $v_i$. The subset of human-made objects is composed of points with large values. The threshold is provided by the GMM and the $k$-means methods. We apply~\eqref{eq:kernel estimator of the Hessian matrix} with $\hat{\mathcal A}$ given by the original labels for the nonground points in Figure~\ref{fig:satellite map city}(b) to the GMM method. We obtain Figure~\ref{fig:Cluster without ground filtering}(a). We find that a better result can be derived if the proposed statistical ground filtering is used in the LIE stage. We present our result in Section~\ref{sec:experiment}.


\section{Experiment}
\label{sec:experiment}

\begin{figure}[tb]
     \centering
     \begin{subfigure}[b]{0.22\textwidth}
         \centering
         \includegraphics[width=\textwidth]{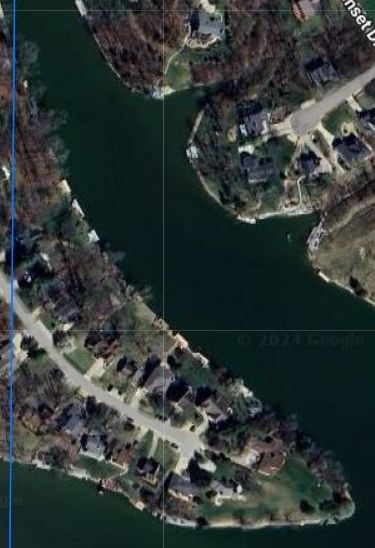}
         \caption{Satelite map}
     \end{subfigure}
     \begin{subfigure}[b]{0.22\textwidth}
         \centering
         \includegraphics[width=\textwidth]{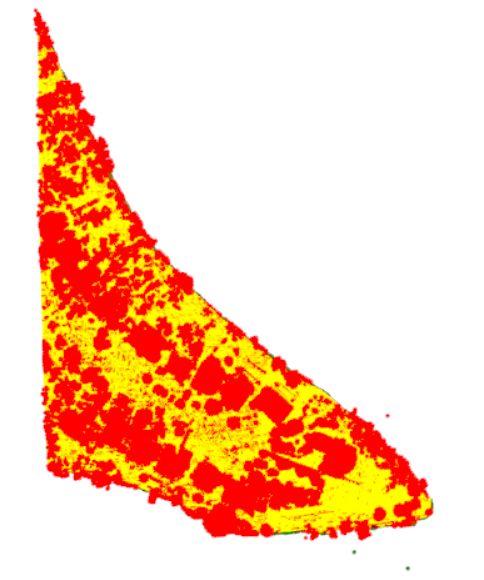} 
         \caption{Original label}
     \end{subfigure}
          \centering
     \begin{subfigure}[b]{0.22\textwidth}
         \centering
         \includegraphics[width=\textwidth]{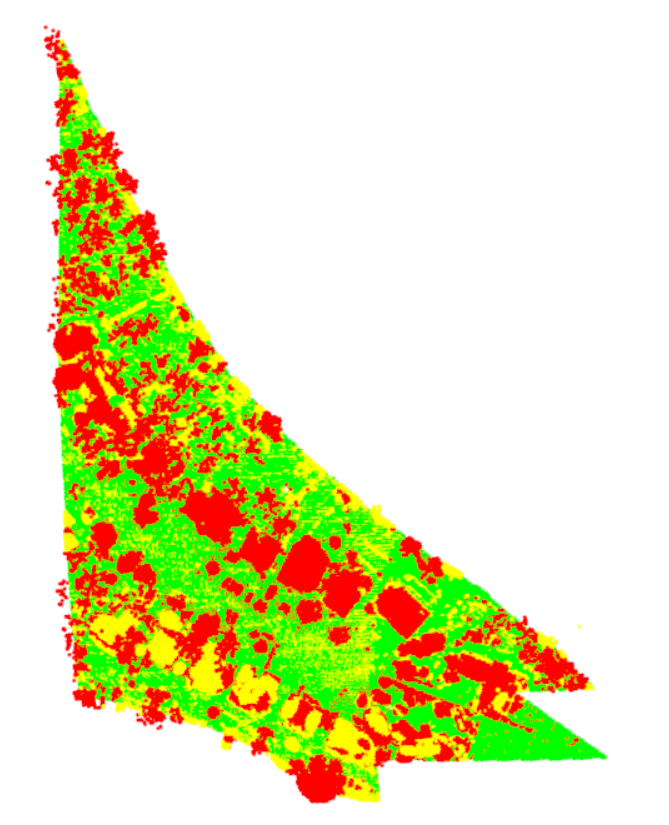}
         \caption{Trees vs human-made objects for nonground points}
     \end{subfigure}
     \begin{subfigure}[b]{0.22\textwidth}
         \centering
         \includegraphics[width=\textwidth]{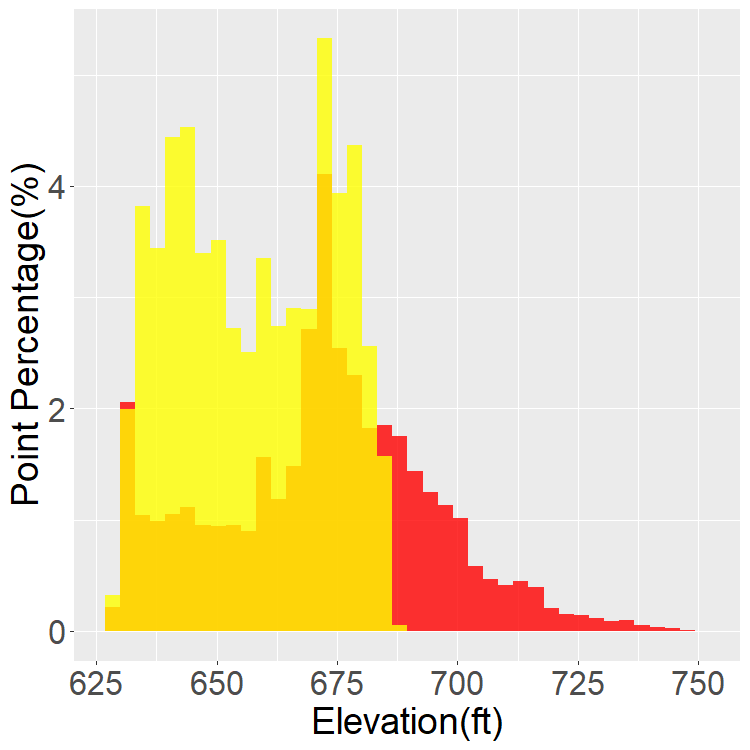} 
         \caption{Histogram for elevations}
     \end{subfigure}
     \caption{\label{fig:satellite map dearborn county} (a) Satellite map for an airborne LiDAR dataset from a region in Dearborn County, Indiana; (b) original labels for {\it ground} (yellow) and {\it nonground} (red) points; (c) Unsupervised ML/AL for human-made objects (yellow) and trees (red), derived by combining our proposed LIE method with the previous GMM method for {\it nonground} points; (d) Histogram showing the percentage of elevations for {\it ground} (yellow) and {\it nonground} (red) points.}
\end{figure}

We applied our method with the comparison to a few previous ones to the real 3D point cloud data downloaded from the same website of Figure~\ref{fig:satellite map city}. We investigated various geological regions in the data. We found that the finding could be represented by two of those. The first was the region displayed in Figure~\ref{fig:satellite map city}. The second was a $844.03\times 1237.66({\rm ft}^2)$ region with a $0.36/{\rm ft}^2$ point density in a lake-side community called Hidden Valley in the Dearborn County, Indiana (Figure~\ref{fig:satellite map dearborn county}). The dataset contained the longitude, latitude, and elevation values of the points with an important associated mark variable for the classification of ground/nonground for points derived by the same slope-based method used in Figure~\ref{fig:satellite map city}. The difference was that the region of Figure~\ref {fig:satellite map city} was flat but the region of Figure ~\ref{fig:satellite map dearborn county} was composed of a few inclined surfaces from the water area to the top of the region. The evidence was identified by the comparison between Figures~\ref{fig:Cluster without ground filtering}(b) and \ref{fig:satellite map dearborn county}(d). We applied our proposed LIE method introduced in Section~\ref{subsec:local information extraction (lie)} to the subset of nonground points claimed by the mark variable (i.e., Figures~\ref{fig:satellite map city}(b) and~\ref{fig:satellite map dearborn county}(b)). We applied the results of the LIE to the GMM for clustering. It partitioned the nonground points into subsets of trees and human-made objects (i.e., Figures~\ref{fig:Cluster without ground filtering}(a) and~\ref{fig:satellite map dearborn county}(c)). The misclassification rate of Figure~\ref{fig:satellite map dearborn county}(c) was much higher than that of Figure~\ref{fig:Cluster without ground filtering}(a). The reason was that the ground surface Figure~\ref{fig:satellite map city}(a) was flat but that of Figure~\ref{fig:satellite map dearborn county}(a) was. 

Our initial study shows that ground filtering is important in human-made object detection. It significantly affects the results of LIE and clustering. The site's slope-based methods do not separate ground and nonground points well, leading to too many mistakes in partitioning the nonground points into subsets of trees and human-made objects. It is necessary to investigate the entire procedure for the research problem. Existing ground filtering methods cannot address the challenge. A new ground filtering method is needed. 

\begin{figure}[tb]
     \centering
     \begin{subfigure}[b]{0.22\textwidth}
         \centering
         \includegraphics[width=\textwidth]{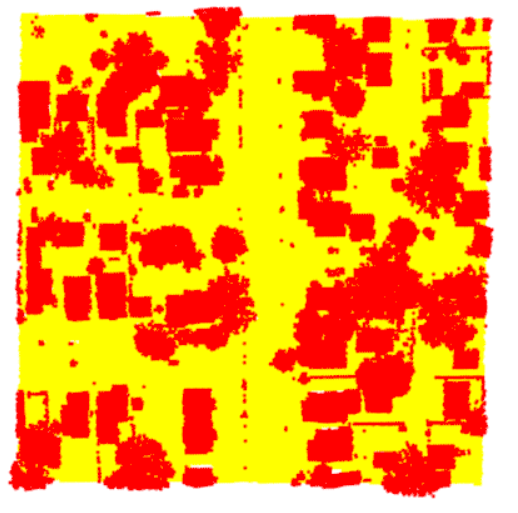}
         \caption{OSR}
     \end{subfigure}
     \begin{subfigure}[b]{0.22\textwidth}
         \centering
         \includegraphics[width=\textwidth]{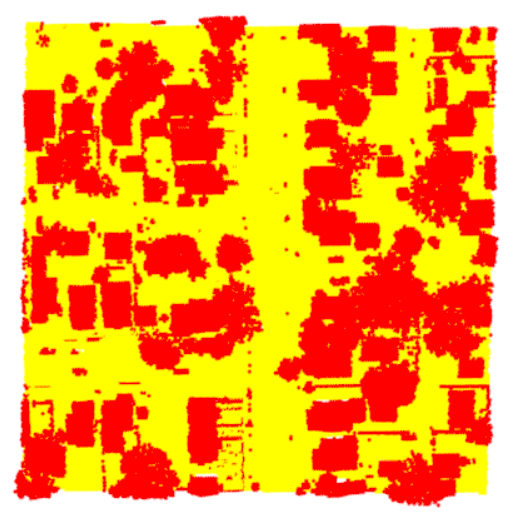} 
         \caption{CSF}
     \end{subfigure}
     \begin{subfigure}[b]{0.22\textwidth}
         \centering
         \includegraphics[width=\textwidth]{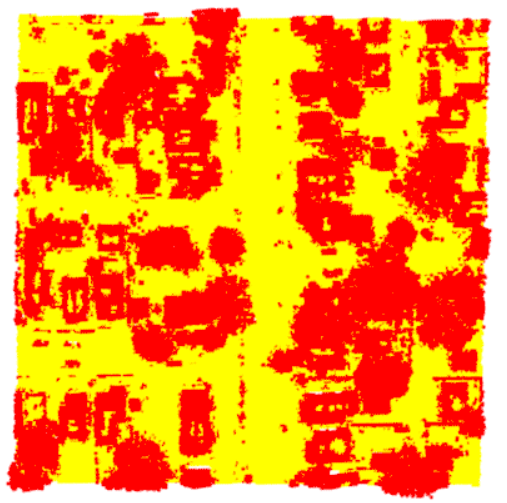} 
         \caption{PMF}
     \end{subfigure}
     \begin{subfigure}[b]{0.22\textwidth}
         \centering
         \includegraphics[width=\textwidth]{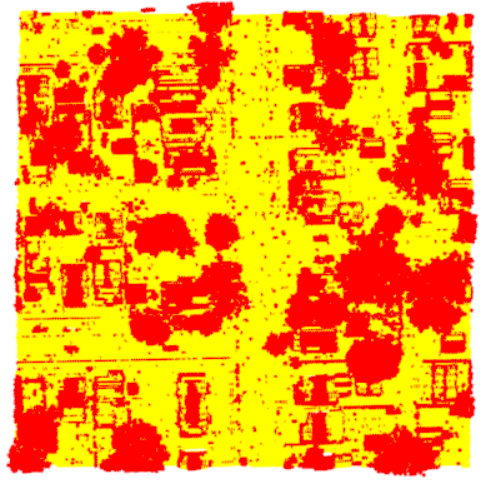} 
         \caption{MCC}
     \end{subfigure}
     \caption{\label{fig:ground filtering 4 methods}ground filtering for Figure~\ref{fig:satellite map city}(a) using our proposed OSR method and three previous state-of-the-art methods, where {\it yellow} represents {\it ground} and {\it red} represent nonground points.}
\end{figure}

The slope-based methods did not work well. We investigated two other categories of ground filtering proposed in the literature. They were the categories of the mathematical morphology-based and surface-based methods reviewed in Section~\ref{sec:related work}. The three well-known state-of-art methods extensively used in applications were the CSF~\cite{zhang2016easy}, the PMF~\cite{zhang2003progressive}, and the MCC~\cite{evans2007multiscale}. We implemented the three methods via the \textsf{lidR} package of \textsf{R}. We found that the CSF, the PMF, and the MCC methods were unreliable for uneven terrains. We then decided to devise our own OSR method in our research. The OSR was fundamentally different from the previous ground filtering methods. Because it is developed under a statistical model, we classify it as a new category of statistical ground filtering methods in this article. We applied the four methods to Figure~\ref{fig:satellite map city}(a) for ground filtering, leading to Figure~\ref{fig:ground filtering 4 methods}. 

We visually compared Figures~\ref{fig:ground filtering 4 methods} with~\ref{fig:satellite map city}(b) for the performance of ground filtering based on the satellite map given by Figure~\ref{fig:satellite map city}(a). The results showed that our OSR was the best. The corresponding classification of the ground/nonground points matched the information from our eyes. The classification of those provided by the CSF was close to that of our OSR. A significant difference was found in the lower middle parts of the figures. Our OSR and the previous CSF were better than the previous PMF, MCC, and sloped-based methods. The PMF misclassified too many points on the top of roofs as ground points. The sloped-based method adopted by the website of the data misclassified too many ground points as nonground points. The situation became worse in the MCC method.

We started the LIE stage after the ground filtering stage was finished. We applied our LIE method to the subsets of nonground points provided by the five methods respectively. We examined various options of the bandwidth vector for~\eqref{eq:kernel estimator of the Hessian matrix} and~\eqref{eq:3D gaussian kernel}. The optimal bandwidth ${\bm h}=(h_x,h_y,h_z)^\top$ used in the formulations depended on point densities. After ground points were removed, the densities of nonground points varied substantially across the study region. Small bandwidth values are unreliable. They cannot be used if point densities are low in a local area. If a larger bandwidth value is used, then a more smooth estimator of the Hessian matrix is induced. Thus, an appropriate choice of ${\bm h}$ is needed. We investigated the optimal bandwidth problem based on this issue. We found that the optimal $h_x=h_y$ could vary from $4$ to $8{\rm ft}$, and the optimal $h_z$ could vary from $6$ to $10{\rm ft}$. We slightly varied the bandwidths around the optimal ${\bm h}$. We did not find significant changes. Therefore, we could implement~\eqref{eq:kernel estimator of the Hessian matrix} and~\eqref{eq:3D gaussian kernel} based on the optimal ${\bm h}$. 


\begin{figure}[tb]
     \centering
     \begin{subfigure}[b]{0.22\textwidth}
         \centering
         \includegraphics[width=\textwidth]{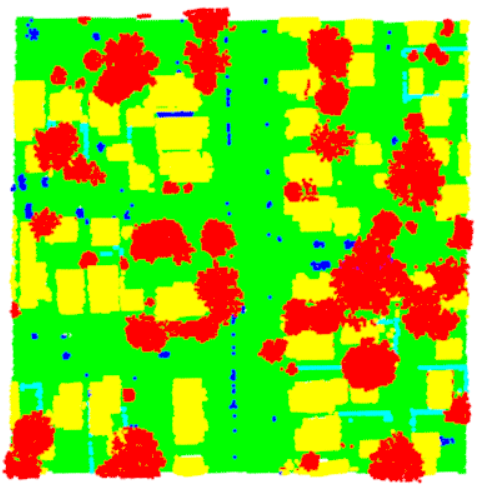}
         \caption{OSR+LIE}
     \end{subfigure}
     \begin{subfigure}[b]{0.22\textwidth}
         \centering
         \includegraphics[width=\textwidth]{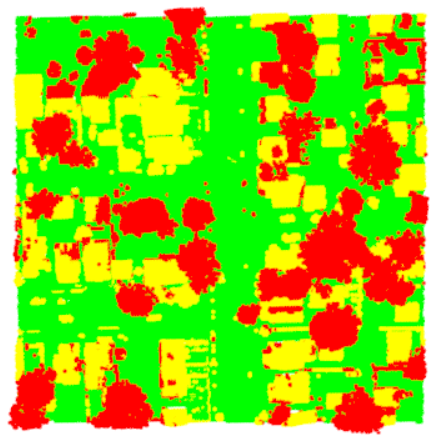} 
         \caption{CSF+LIE}
     \end{subfigure}
     \begin{subfigure}[b]{0.22\textwidth}
         \centering
         \includegraphics[width=\textwidth]{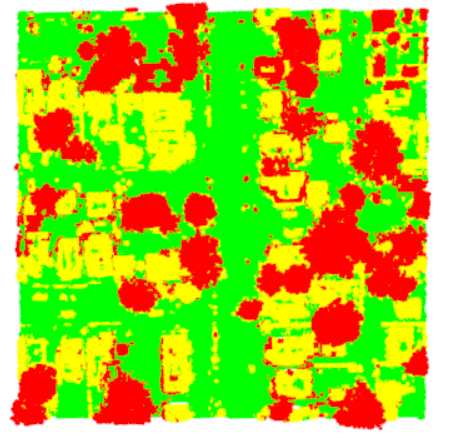} 
         \caption{PMF+LIE}
     \end{subfigure}
     \begin{subfigure}[b]{0.22\textwidth}
         \centering
         \includegraphics[width=\textwidth]{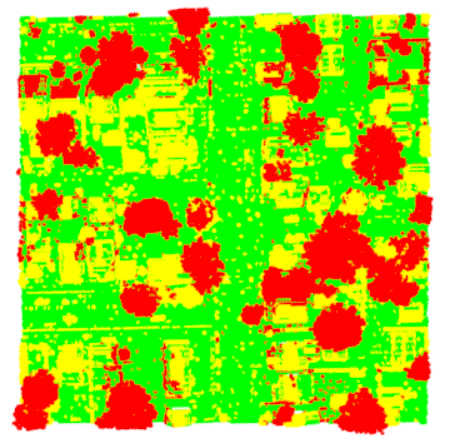} 
         \caption{MCC+LIE}
     \end{subfigure}
     \caption{\label{fig:clustering 4 methods west lafayette region}Trees and human-made objects identified by the GMM clustering, based on the combination of our LIE method with our OSR, and the previous CSF, PMF, and MCC methods, respectively, for the nonground points in Figure~\ref{fig:ground filtering 4 methods}. {\it Green} represents ground points, {\it red} represents trees, and {\it yellow} represents human-made objects.}
\end{figure}

We estimated the Hessian matrix of the first-order intensity for every nonground point claimed by the five ground filtering methods. The Hessian matrix was $3\times 3$ and symmetric. It had $3$ real eigenvalues. The least eigenvalue of the Hessian matrix for points from the human-made objects was close to $0$ but not for points from the trees. We examined the negative log of the proportion of the squares of the least eigenvalue, leading to $v_i$ given by~\eqref{eq:choice of local variables eigenvalue}. We found that $v_i$ better interpreted the difference between points from the trees and human-made objects. 

In the clustering stage, we treated $v_i$ as the key local feature. We also considered a mark variable called intensity for the strength of laser pulses. We incorporated those into the GMM and $k$-means. We found that the GMM was more reliable than the $k$-means. We discarded the $k$-means and adopted the GMM, leading to Figure~\ref{fig:clustering 4 methods west lafayette region}. 

A common challenge in unsupervised ML is the number of clusters. It can be easily addressed because the purpose is to partition the subset of the nonground points into two subsets. We chose two clusters in the GMM. We also considered three clusters to evaluate the impact of the number of clusters. We found that the GMM with two clusters was the best in the combinations of our LIE with the previous CSF, PMF, MCC, and sloped-based methods. New findings appeared in the combination of our LIE and OSR. The GMM further partitioned human-made objects into houses ({\it yellow}) and vehicles ({\it blue}), leading to Figure~\ref{fig:clustering 4 methods west lafayette region}(a).  

We visually compared the results of the GMM with the satellite map for ground/nonground points. The best was the combination of our LIE with our OSR. A significant difference was identified at the upper right corner. The house was misclassified as trees in the combinations of our LIE with the previous PMF and the sloped-based methods. The combinations of our LIE and the previous CSF and MCC methods found the house, but they misclassified another house below. Overall, the pattern provided by the combination of our LIE and our OSR was clearer than those by other combinations. 


\begin{figure}[tb]
     \centering
     \begin{subfigure}[b]{0.22\textwidth}
         \centering
         \includegraphics[width=\textwidth]{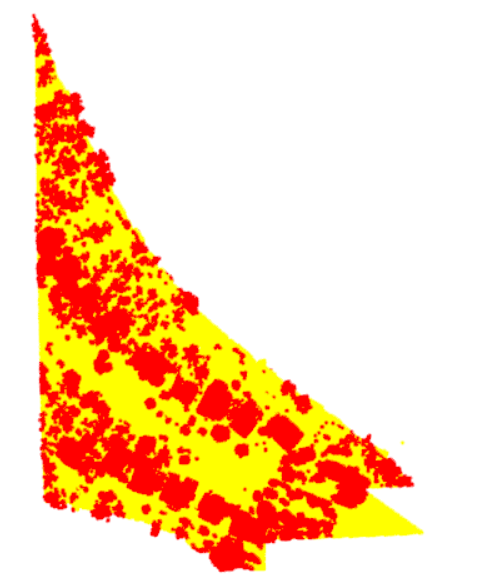}
         \caption{OSR}
     \end{subfigure}
     \begin{subfigure}[b]{0.22\textwidth}
         \centering
         \includegraphics[width=\textwidth]{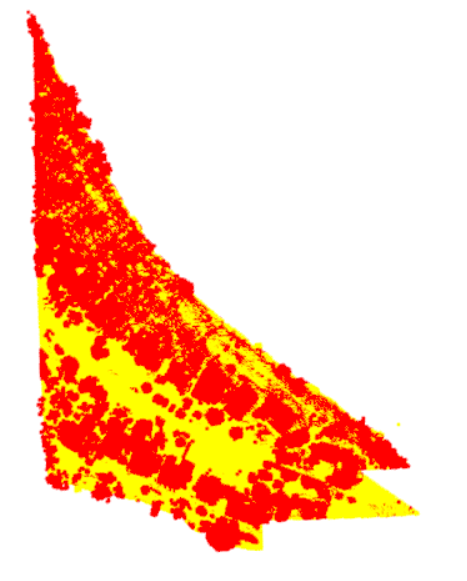} 
         \caption{CSF}
     \end{subfigure}
     \begin{subfigure}[b]{0.22\textwidth}
         \centering
         \includegraphics[width=\textwidth]{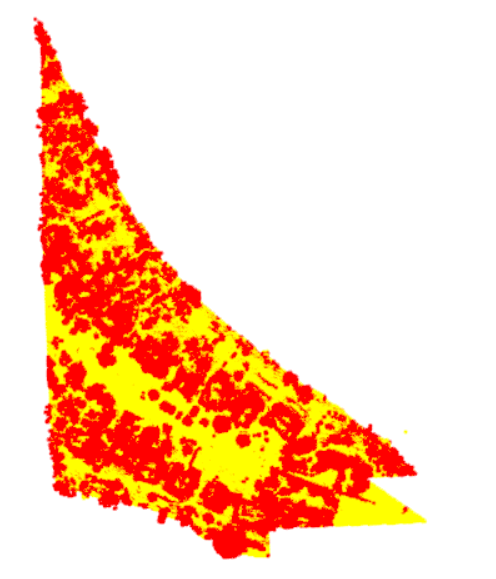} 
         \caption{PMF}
     \end{subfigure}
     \begin{subfigure}[b]{0.22\textwidth}
         \centering
         \includegraphics[width=\textwidth]{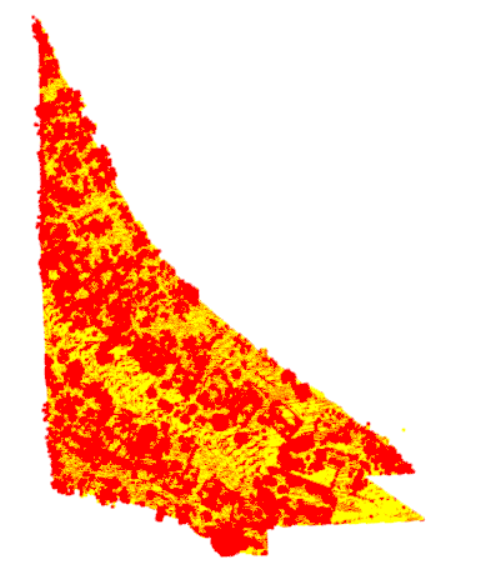} 
         \caption{MCC}
     \end{subfigure}
     \caption{\label{fig:ground filtering 4 methods hidden valley}Ground filtering for Figure~\ref{fig:satellite map dearborn county}(a) using our proposed OSR method and the previous CSF, PMF, and MCC methods, where {\it yellow} represents {\it ground} points and {\it red} represents nonground points.}
\end{figure}

We next moved our interest to Figure~\ref{fig:satellite map dearborn county}. The goal was to investigate the impacts of various types of terrains on our research problem. The ground surface was flat in Figure~\ref{fig:satellite map city}(a) but not in Figure~\ref{fig:satellite map dearborn county}(a). The application for Figure~\ref{fig:satellite map dearborn county}(a) can reflect the reliability of a method to the region topography. 

The website of the Lidar data used a slope-based method for ground filtering, leading to Figure~\ref{fig:satellite map dearborn county}(b). We examined the elevations of the points (i.e., Figure~\ref{fig:satellite map dearborn county}(d)) and found that the sloped-based method classified too many ground as nonground points at the border of the lake. Similar issues also happened in the CSF and MCC methods (Figure~\ref{fig:ground filtering 4 methods hidden valley}). The PMF method worked better than the sloped-based, CSF, and MCC methods. Our proposed OSR provided the best result for ground filtering. 

\begin{figure}[tb]
     \centering
     \begin{subfigure}[b]{0.22\textwidth}
         \centering
         \includegraphics[width=\textwidth]{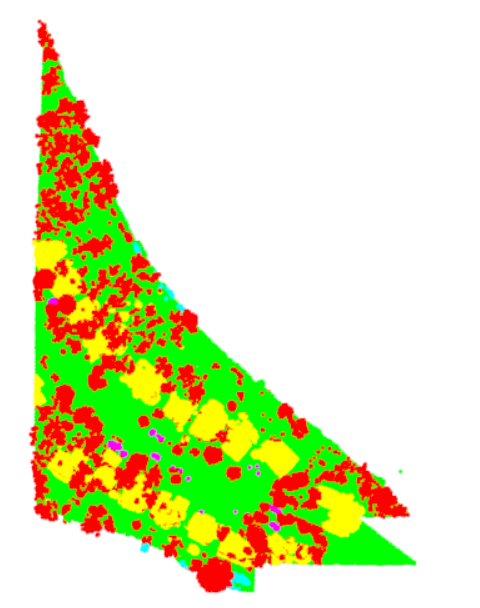}
         \caption{OSR+LIE}
     \end{subfigure}
     \begin{subfigure}[b]{0.22\textwidth}
         \centering
         \includegraphics[width=\textwidth]{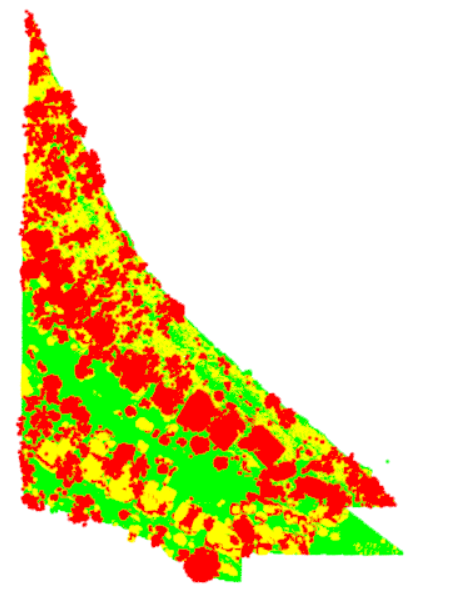} 
         \caption{CSF+LIE}
     \end{subfigure}
     \begin{subfigure}[b]{0.22\textwidth}
         \centering
         \includegraphics[width=\textwidth]{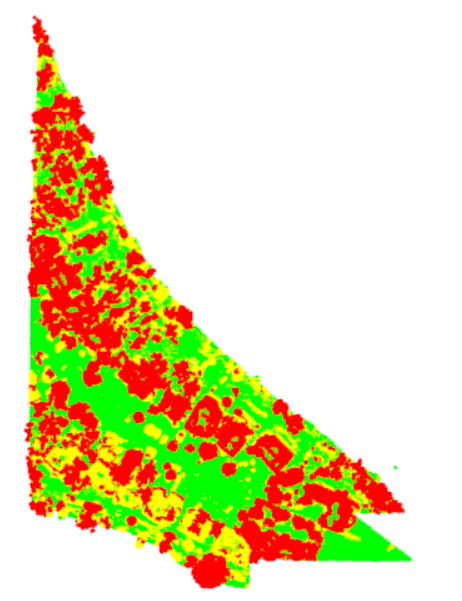} 
         \caption{PMF+LIE}
     \end{subfigure}
     \begin{subfigure}[b]{0.22\textwidth}
         \centering
         \includegraphics[width=\textwidth]{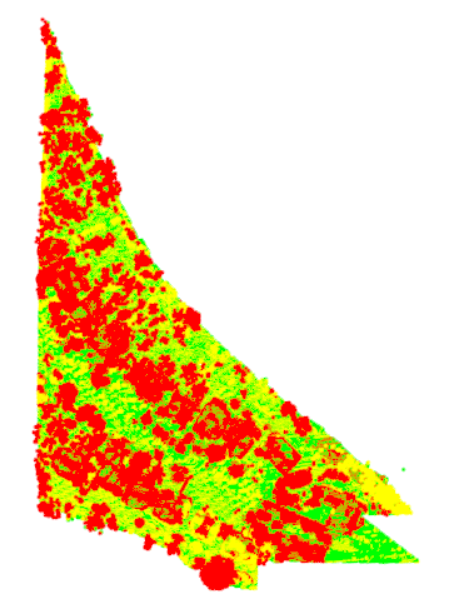} 
         \caption{MCC+LIE}
     \end{subfigure}
     \caption{\label{fig:clustering 4 methods hidden valley}Trees and human-made objects identified by GMM clustering, based on the combination of our LIE with our OSR method, and the previous CSF, PMF, and MCC methods, respectively, for the nonground points in Figure~\ref{fig:ground filtering 4 methods hidden valley}. {\it Green} represents ground points, {\it red} represents trees, and {\it yellow} represents human-made objects.}
\end{figure}

Similarly, we applied our method for the LIE stage to the subset of nonground points provided by the separation stage. We also investigated the optimal bandwidth ${\bm h}$ for~\eqref{eq:kernel estimator of the Hessian matrix} and~\eqref{eq:3D gaussian kernel}. We chose larger bandwidth values in local areas with low nonground point densities or vice versa. We found that the optimal $h_x=h_y$ could vary from $4$ to $8{\rm ft}$, and the optimal $h_z$ could vary from $4$ to $15{\rm ft}$. We also slightly varied the bandwidths around the optimal ${\bm h}$. We did not find significant changes either. We also estimated the Hessian matrix of the first-order intensity for nonground points claimed by the five ground filtering methods. We incorporated $v_i$ given by~\eqref{eq:choice of local variables eigenvalue} into the GMM with two or three clusters. We found that the GMM with two clusters better interpreted the results, leading to Figures~\ref{fig:satellite map dearborn county}(c) and~\ref{fig:clustering 4 methods hidden valley}. 

We next visually compared the results. The combination of our LIE with OSR found the lower and the upper rows of the houses. The combination of our LIE with the MCC did not find any row of the houses. The combinations of our LIE with the sloped-based, CSF, and PMF found the lower but not upper row of the houses. They misclassified many ground points as human-made objects. The proposed OSR was the best. The previous MCC was the worst. The previous sloped-based, CSF, and PMF were close to each other. 

The topography of terrains significantly affects the performance of our method. The region in Figure~\ref{fig:satellite map city}(a) is flat but the region in~\ref{fig:satellite map dearborn county}(a) is not. The previous ground filtering methods work well in the former but not in the latter. Our proposed OSR works well in both. The proposed LIE stage works well if ground/nonground points are well separated but not otherwise. It is appropriate to use our proposed OSR method for ground filtering. 

The unsupervised learning method can detect and locate human-made objects. Existence is concluded if the subset of human-made objects is nonempty. Location can be found by the longitude, latitude, and elevation of the points claimed as human-made objects. The proposed method is effective even if the human-made objects are hidden under the trees. The colors of the human-made objects are not an issue. The shape of the surface is the main property of our method.

\section{Conclusion}
\label{sec:conclusion}
An airborne 3D point cloud comprises a collection of points derived from a variety of objects, including both ground and non-ground elements. Identifying human-made structures within a geological region is a significant focus in many real-world applications. Given the lack of ground truth data, unsupervised learning is more suitable than supervised learning for processing airborne 3D point cloud data. The three stages proposed in our method—ground filtering, LIE, and clustering—constitute a general framework for developing unsupervised learning techniques. These stages are independently developed and can be integrated with various methods as needed. Ground filtering is particularly crucial, and the proposed OSR technique is effective in this context. OSR represents a novel approach within the emerging category of statistical ground filtering, an area that has not been extensively explored in the literature. The potential of OSR to significantly influence the development of machine learning and artificial intelligence methods for 3D point cloud data remains a promising avenue for future research.


\bibliographystyle{refs/IEEEtran}

\begin{thebibliography}{10}
\providecommand{\url}[1]{#1}
\csname url@samestyle\endcsname
\providecommand{\newblock}{\relax}
\providecommand{\bibinfo}[2]{#2}
\providecommand{\BIBentrySTDinterwordspacing}{\spaceskip=0pt\relax}
\providecommand{\BIBentryALTinterwordstretchfactor}{4}
\providecommand{\BIBentryALTinterwordspacing}{\spaceskip=\fontdimen2\font plus
\BIBentryALTinterwordstretchfactor\fontdimen3\font minus \fontdimen4\font\relax}
\providecommand{\BIBforeignlanguage}[2]{{%
\expandafter\ifx\csname l@#1\endcsname\relax
\typeout{** WARNING: IEEEtran.bst: No hyphenation pattern has been}%
\typeout{** loaded for the language `#1'. Using the pattern for}%
\typeout{** the default language instead.}%
\else
\language=\csname l@#1\endcsname
\fi
#2}}
\providecommand{\BIBdecl}{\relax}
\BIBdecl

\bibitem{shan2018topographic}
J.~Shan and C.~K. Toth, \emph{Topographic laser ranging and scanning: principles and processing}.\hskip 1em plus 0.5em minus 0.4em\relax CRC press, 2018.

\bibitem{qi2017pointnet}
C.~R. Qi, H.~Su, K.~Mo, and L.~J. Guibas, ``Pointnet: Deep learning on point sets for 3d classification and segmentation,'' in \emph{Proceedings of the IEEE conference on computer vision and pattern recognition}, 2017, pp. 652--660.

\bibitem{ogata1998space}
Y.~Ogata, ``Space-time point-process models for earthquake occurrences,'' \emph{Annals of the Institute of Statistical Mathematics}, vol.~50, pp. 379--402, 1998.

\bibitem{zhang2017independence}
T.~Zhang, ``On independence and separability between points and marks of marked point processes,'' \emph{Statistica Sinica}, pp. 207--227, 2017.

\bibitem{zhuang2002stochastic}
J.~Zhuang, Y.~Ogata, and D.~Vere-Jones, ``Stochastic declustering of space-time earthquake occurrences,'' \emph{Journal of the American Statistical Association}, vol.~97, no. 458, pp. 369--380, 2002.

\bibitem{peng2005space}
R.~D. Peng, F.~P. Schoenberg, and J.~A. Woods, ``A space--time conditional intensity model for evaluating a wildfire hazard index,'' \emph{Journal of the American Statistical Association}, vol. 100, no. 469, pp. 26--35, 2005.

\bibitem{schoenberg2004testing}
F.~P. Schoenberg, ``Testing separability in spatial-temporal marked point processes,'' \emph{Biometrics}, pp. 471--481, 2004.

\bibitem{zhang2014kolmogorov}
T.~Zhang, ``A kolmogorov-smirnov type test for independence between marks and points of marked point processes,'' \emph{Electronic Journal of Statistics}, pp. 2557--2584, 2014.

\bibitem{daley2003introduction}
D.~J. Daley, D.~Vere-Jones \emph{et~al.}, \emph{An introduction to the theory of point processes: volume I: elementary theory and methods}.\hskip 1em plus 0.5em minus 0.4em\relax Springer, 2003.

\bibitem{zhang2017gradient}
T.~Zhang and Y.-N. Huang, ``Gradient angle-based analysis for spatiotemporal point processes,'' \emph{Electronic Journal of Statistics}, pp. 4424--4451, 2017.

\bibitem{vosselman2000slope}
G.~Vosselman, ``Slope based filtering of laser altimetry data,'' \emph{International archives of photogrammetry and remote sensing}, vol.~33, no. B3/2; PART 3, pp. 935--942, 2000.

\bibitem{shan2005urban}
J.~Shan and S.~Aparajithan, ``Urban dem generation from raw lidar data,'' \emph{Photogrammetric Engineering \& Remote Sensing}, vol.~71, no.~2, pp. 217--226, 2005.

\bibitem{meng2009multi}
X.~Meng, L.~Wang, J.~L. Silv{\'a}n-C{\'a}rdenas, and N.~Currit, ``A multi-directional ground filtering algorithm for airborne lidar,'' \emph{ISPRS journal of Photogrammetry and Remote Sensing}, vol.~64, no.~1, pp. 117--124, 2009.

\bibitem{sithole2001filtering}
G.~Sithole and G.~Vosselman, ``Filtering of laser altimetry data using a slope adaptive filter,'' \emph{International Archives of Photogrammetry Remote Sensing and Spatial Information Sciences}, vol.~34, no. 3/W4, pp. 203--210, 2001.

\bibitem{streutker2011slope}
D.~R. Streutker, N.~F. Glenn, and R.~Shrestha, ``A slope-based method for matching elevation surfaces,'' \emph{Photogrammetric Engineering \& Remote Sensing}, vol.~77, no.~7, pp. 743--750, 2011.

\bibitem{susaki2012adaptive}
J.~Susaki, ``Adaptive slope filtering of airborne lidar data in urban areas for digital terrain model (dtm) generation,'' \emph{Remote Sensing}, vol.~4, no.~6, pp. 1804--1819, 2012.

\bibitem{wan2018simple}
P.~Wan, W.~Zhang, A.~K. Skidmore, J.~Qi, X.~Jin, G.~Yan, and T.~Wang, ``A simple terrain relief index for tuning slope-related parameters of lidar ground filtering algorithms,'' \emph{ISPRS journal of photogrammetry and remote sensing}, vol. 143, pp. 181--190, 2018.

\bibitem{liu2008airborne}
X.~Liu, ``Airborne lidar for dem generation: some critical issues,'' \emph{Progress in physical geography}, vol.~32, no.~1, pp. 31--49, 2008.

\bibitem{sithole2005filtering}
G.~Sithole and G.~Vosselman, ``Filtering of airborne laser scanner data based on segmented point clouds,'' in \emph{ISPRS Workshop Laser Scanning 2005}.\hskip 1em plus 0.5em minus 0.4em\relax International Institute for Geo-Information Science and Earth Observation, 2005, pp. 66--71.

\bibitem{zhang2003progressive}
K.~Zhang, S.-C. Chen, D.~Whitman, M.-L. Shyu, J.~Yan, and C.~Zhang, ``A progressive morphological filter for removing nonground measurements from airborne lidar data,'' \emph{IEEE transactions on geoscience and remote sensing}, vol.~41, no.~4, pp. 872--882, 2003.

\bibitem{chen2007filtering}
Q.~Chen, P.~Gong, D.~Baldocchi, and G.~Xie, ``Filtering airborne laser scanning data with morphological methods,'' \emph{Photogrammetric Engineering \& Remote Sensing}, vol.~73, no.~2, pp. 175--185, 2007.

\bibitem{mongus2012parameter}
D.~Mongus and B.~{\v{Z}}alik, ``Parameter-free ground filtering of lidar data for automatic dtm generation,'' \emph{ISPRS Journal of Photogrammetry and Remote Sensing}, vol.~67, pp. 1--12, 2012.

\bibitem{axelsson2000generation}
P.~Axelsson, ``Dem generation from laser scanner data using adaptive tin models,'' \emph{International archives of photogrammetry and remote sensing}, vol.~33, no.~4, pp. 110--117, 2000.

\bibitem{kraus1998determination}
K.~Kraus and N.~Pfeifer, ``Determination of terrain models in wooded areas with airborne laser scanner data,'' \emph{ISPRS Journal of Photogrammetry and remote Sensing}, vol.~53, no.~4, pp. 193--203, 1998.

\bibitem{pedregosa2011scikit}
F.~Pedregosa, G.~Varoquaux, A.~Gramfort, V.~Michel, B.~Thirion, O.~Grisel, M.~Blondel, P.~Prettenhofer, R.~Weiss, V.~Dubourg \emph{et~al.}, ``Scikit-learn: Machine learning in python,'' \emph{Journal of machine learning research}, vol.~12, no. Oct, pp. 2825--2830, 2011.

\bibitem{chen2013multiresolution}
C.~Chen, Y.~Li, W.~Li, and H.~Dai, ``A multiresolution hierarchical classification algorithm for filtering airborne lidar data,'' \emph{ISPRS journal of photogrammetry and remote sensing}, vol.~82, pp. 1--9, 2013.

\bibitem{su2015new}
W.~Su, Z.~Sun, R.~Zhong, J.~Huang, M.~Li, J.~Zhu, K.~Zhang, H.~Wu, and D.~Zhu, ``A new hierarchical moving curve-fitting algorithm for filtering lidar data for automatic dtm generation,'' \emph{International Journal of Remote Sensing}, vol.~36, no.~14, pp. 3616--3635, 2015.

\bibitem{hui2016improved}
Z.~Hui, Y.~Hu, Y.~Z. Yevenyo, and X.~Yu, ``An improved morphological algorithm for filtering airborne lidar point cloud based on multi-level kriging interpolation,'' \emph{Remote Sensing}, vol.~8, no.~1, p.~35, 2016.

\bibitem{elmqvist2000automatic}
M.~Elmqvist, ``Automatic ground modeling using laser radar data,'' \emph{Master's thesis LiTH-ISY-EX-3061, Linkoping University}, 2000.

\bibitem{guan2014generation}
H.~Guan, J.~Li, Y.~Yu, L.~Zhong, and Z.~Ji, ``Dem generation from lidar data in wooded mountain areas by cross-section-plane analysis,'' \emph{International Journal of Remote Sensing}, vol.~35, no.~3, pp. 927--948, 2014.

\bibitem{zhang2016easy}
W.~Zhang, J.~Qi, P.~Wan, H.~Wang, D.~Xie, X.~Wang, and G.~Yan, ``An easy-to-use airborne lidar data filtering method based on cloth simulation,'' \emph{Remote sensing}, vol.~8, no.~6, p. 501, 2016.

\bibitem{evans2007multiscale}
J.~S. Evans and A.~T. Hudak, ``A multiscale curvature algorithm for classifying discrete return lidar in forested environments,'' \emph{IEEE Transactions on Geoscience and Remote Sensing}, vol.~45, no.~4, pp. 1029--1038, 2007.

\bibitem{diggle2013statistical}
P.~J. Diggle, \emph{Statistical analysis of spatial and spatio-temporal point patterns}.\hskip 1em plus 0.5em minus 0.4em\relax CRC press, 2013.

\bibitem{ripley1976second}
B.~D. Ripley, ``The second-order analysis of stationary point processes,'' \emph{Journal of applied probability}, vol.~13, no.~2, pp. 255--266, 1976.

\bibitem{besag1977contribution}
J.~Besag, ``Contribution to the discussion on dr ripley's paper,'' \emph{JR Stat Soc B}, vol.~39, pp. 193--195, 1977.

\bibitem{stoyan1996estimating}
D.~Stoyan and H.~Stoyan, ``Estimating pair correlation functions of planar cluster processes,'' \emph{Biometrical Journal}, vol.~38, no.~3, pp. 259--271, 1996.

\bibitem{moller2007modern}
J.~M{\o}ller and R.~P. Waagepetersen, ``Modern statistics for spatial point processes,'' \emph{Scandinavian Journal of Statistics}, vol.~34, no.~4, pp. 643--684, 2007.

\bibitem{baddeley2000non}
A.~J. Baddeley, J.~M{\o}ller, and R.~Waagepetersen, ``Non-and semi-parametric estimation of interaction in inhomogeneous point patterns,'' \emph{Statistica Neerlandica}, vol.~54, no.~3, pp. 329--350, 2000.

\bibitem{zhang2019substationarity}
T.~Zhang and J.~Mateu, ``Substationarity for spatial point processes,'' \emph{Journal of Multivariate Analysis}, vol. 171, pp. 22--36, 2019.

\bibitem{loffler2021optimality}
M.~L{\"o}ffler, A.~Y. Zhang, and H.~H. Zhou, ``Optimality of spectral clustering in the gaussian mixture model,'' \emph{The Annals of Statistics}, vol.~49, no.~5, pp. 2506--2530, 2021.

\bibitem{huang2023improved}
H.~Huang, Z.~Tang, T.~Zhang, and B.~Yang, ``Improved clustering using nice initialization,'' in \emph{GLOBECOM 2023-2023 IEEE Global Communications Conference}.\hskip 1em plus 0.5em minus 0.4em\relax IEEE, 2023, pp. 320--325.

\bibitem{donoho2004higher}
D.~Donoho and J.~Jin, ``Higher criticism for detecting sparse heterogeneous mixtures,'' \emph{The Annals of Statistics}, vol.~32, no.~3, pp. 962--994, 2004.

\end{thebibliography}
 \newcommand{\noop}[1]{}

\end{document}